\documentclass[letterpaper]{article} 
\usepackage[submission]{aaai23}  
\usepackage{times}  
\usepackage{helvet}  
\usepackage{courier}  
\usepackage[hyphens]{url}  
\usepackage{graphicx} 
\usepackage{natbib}  
\usepackage{caption} 
\usepackage{algorithm}
\usepackage{algorithmic}
\usepackage{comment}
\usepackage{color}
\usepackage{graphicx}
\usepackage{amsmath}
\usepackage{booktabs}
\usepackage{color}
\usepackage{bm}
\usepackage{algorithm}
\usepackage{algorithmic}

\usepackage{amsmath}
\usepackage{amssymb}\usepackage{multirow}
\usepackage{enumitem}
\usepackage{caption}
\usepackage{lipsum}
\usepackage[flushleft]{threeparttable}
\usepackage{mathtools} 
\DeclarePairedDelimiter{\floor}{\lfloor}{\rfloor}
\usepackage{tikz}
\usetikzlibrary{arrows.meta}

\newcommand\scalemath[2]{\scalebox{#1}{\mbox{\ensuremath{\displaystyle #2}}}}

\usepackage{verbatim}

\usepackage{newfloat}
\usepackage{listings}
\DeclareCaptionStyle{ruled}{labelfont=normalfont,labelsep=colon,strut=off} 
\floatstyle{ruled}
\newfloat{listing}{tb}{lst}{}
\floatname{listing}{Listing}
\title{Low Latency Conversion of Artificial Neural Network Models to \\
Rate-encoded Spiking Neural Networks}
\author{
    Zhanglu Yan\textsuperscript{\rm 1}, Jun Zhou\textsuperscript{\rm 2}, Weng-Fai Wong\textsuperscript{\rm 1}
}
\affiliations{
    \textsuperscript{\rm 1}Department of Computer Science\\
  National University of Singapore\\
  Singapore, 117417 \\
  
  \textsuperscript{\rm 2}Agency for Science, Technology and Research\\
  Institute of High Performance Computing \\
  Singapore, 138632 \\
  zhangluyan@comp.nus.edu.sg, zhoujun@ihpc.a-star.edu.sg, wongwf@nus.edu.sg

}
\begin{document}

\maketitle

\begin{abstract}

Spiking neural networks (SNNs) are well suited for resource-constrained applications as they do not need expensive multipliers. In a typical rate-encoded SNN, a series of binary spikes within a globally fixed time window is used to fire the neurons. The maximum number of spikes in this time window is also the latency of the network in performing a single inference, as well as determines the overall energy efficiency of the model. The aim of this paper is to reduce this while maintaining accuracy when converting ANNs to their equivalent SNNs. 
The state-of-the-art conversion schemes yield SNNs with accuracies comparable with ANNs only for large window sizes.  In this paper, we start with understanding the information loss when converting from pre-existing ANN models to standard rate-encoded SNN models. From these insights, we propose a suite of novel techniques that together mitigate the information lost in the conversion, and achieve state-of-art SNN accuracies along with very low latency. Our method achieved a Top-1 SNN accuracy of 98.73\% (1 time step) on the MNIST dataset, 76.38\% (8 time steps) on the CIFAR-100 dataset, and 93.71\% (8 time steps) on the CIFAR-10 dataset. On ImageNet, an SNN accuracy of 75.35\%/79.16\% was achieved with 100/200 time steps. 
  
\end{abstract}

\section{Introduction}
\label{Introduction}

Unlike traditional artificial neural networks (ANNs), 
{\em spiking neural networks} (SNNs) use binary spikes as activations between layers instead of floating-point numbers. This leads to a significant reduction in energy consumption because expensive (floating point) multiplications are replaced by (fixed point) additions, making SNNs suitable for resource-constrained applications, especially if custom hardware is built. However, the non-differentiable binary-based activations used in their {\em integrate-and-fire} model (IF) make SNNs difficult to train. There are two popular ways to solve the problem: direct training including the {\em spike-timing-dependent plasticity} (STDP)~\cite{stdp} class of algorithms that is specific to SNNs~\cite{taher19}, and {\em ANN-to-SNN conversion}~\cite{rueckauer2016theory}. They solve the non-differentiable and discontinuous issue of SNN in different ways. Direct training methods have the advantage of being simpler and less resource-demanding. However, thus far, they yield accuracies that are lower than the state-of-the-art ANNs and are very sensitive to the hyper parameters~\cite{kheradpisheh2020temporal,deng2022temporal} that require significant tuning. {\em ANN-to-SNN conversion} converts ANNs to SNNs by first training an ANN and then transferring the weights to an equivalent SNN.  It yields higher accuracy with deeper network structures. This is also convenient in use-cases where, for example, if the user already has an ANN that fits certain hardware or other constraints~\cite{murat2021exploring,yan2022eeg}, then obtaining a more efficient SNN becomes easy. Furthermore, starting with an ANN comes with all the benefits of developing in mature ANN development ecosystems such as Tensorflow or Pytorch.

However, the state of the art for such ANN-converted SNNs often requires a large time window size (which is also equivalent to the {\em latency} of a single inference), especially if a low accuracy loss in the conversion process is desired. In theory, an infinite window size would make the SNN equivalent to its ANN twin. However, besides making a single inference slower, the large size of the time windows (typically in the range of $2000-2500$~\cite{sengupta2019going,han2020rmp}) also leads to higher energy consumption, potentially erasing the energy advantage rate-encoded SNNs have over their ANN counterparts. Several methods such as a hybrid training that combines the conversion with spike-based backpropagation~\cite{rathi2020enabling}, a new residual membrane potential (RMP) spiking neuron ~\cite{han2020rmp} and spiking 
backpropagation~\cite{lee2019enabling} have been proposed to reduce the time window size by $10\times$ to $25\times$ compared to direct conversion. However, hundreds of time steps are still needed to achieve SNN accuracies that are comparable with their ANN counterparts on complex datasets. For example, running on the CIFAR-100 dataset, hybrid training~\cite{rathi2020enabling} and RMP~\cite{han2020rmp} achieved an SNN accuracy of 67.87\% and 68.34\% with a latency of 125 and 250, respectively. Furthermore, if spike-based backpropagation is used for latency reduction, integrated gradients that are proportional to the size of the time window are required in the backward pass. This significantly increases the complexity of the training. Recently, Bu {\em et al.}~\cite{bu2022optimized} reduced the time step to 8 and achieved an SNN accuracy of 60.49\%, but the accuracy drop reported was 3.61\%. 

In this paper, we will focus on how to reduce the time window sizes further while maintaining accuracies comparable with the ANN counterparts, as well as simplifying the computation involved. In particular, we proposed a {\em value-range} (VR) encoding method that exactly matches one ANN with its twin SNN, an {\em input channel extension} (ICE) to augment the input data when time step is small and a novel {\em averaging integrate-and-fire spike generation} (ASG) model which increases the robustness of low latency SNNs. Using these, we achieved an SNN accuracy of 67.09\% (76.04\%) on the CIFAR-100 dataset with a latency of only 2 (8) time steps with a near-zero accuracy drop.\footnote{https://anonymous.4open.science/r/aaai23-low-latency-snn-3F21/README.md}


Our ASG model is a variant of the standard IF model that needs to store all the membrane potential at different time steps to increase the robustness of low latency SNN. This may not be easy to implement especially in real-time tasks or when hardware resources are limited. Hence, we further propose a new {\em channel-wise threshold training} (CTT) method that trains a standard IF model SNN alongside the ASG model. The resultant standard SNN obtained in matters of minutes also has high accuracy and low latency. On a VGG model, after threshold training for SNNs, we achieve an SNN accuracy of 66.34\% (72.60\%) on the CIFAR-100 dataset with a latency of only 16 (32) time steps, 33.1\% (12.77\%) higher than a standard IF model SNN without threshold training.

\section{Background}
\label{sec:background}

ANN neurons pass data (activations) as floating-point or fixed-point numbers. SNN neurons use {\em spike trains} consisting of sequences of binary numbers (0 and 1) that are generated by some firing rule such as {\em integrate and fire} (IF) model~\cite{liu2001spike}. Spike generation is governed by a preset voltage threshold. A spike is fired when the accumulated membrane potential exceeds some preset threshold, resetting the membrane potential. 

Spike trains have a globally fixed length, $T$. All spike trains in an SNN are marched synchronously, in $T$ {\em time steps}.  Hence, $T$ is also called a {\em time window}. There is essentially a synchronous pipeline of spikes through the SNN that outputs one inference result every $T$ times. There are two popular variants of SNNs, namely {\em rate-encoded} and the {\em temporal coding} SNNs. The train of spikes sent between neurons in both is partitioned into {\em time windows} that are globally synchronized. This paper focuses on rate-encoded SNNs as they are more mature, being capable of handling more complex models~\cite{guo2021neural}.

In a {\em rate-encoded} SNN with a time window size of $T$, any (activation) value transmitted between two neurons is a sequence $s = \{s_0, s_1, \ldots, s_{T-1}\}$ , where $s_i$ refers to a binary spike at {\em time step} $i$. Let $|s|$ be the number of ones in $s$. The numbers that can be represented would therefore be from the set of values consisting of 0 (no spike at all in the time window), $1/T$, $2/T$, till $T/T = 1$. Also, in our notations below, we will use `$s^l_j(t)$' to denote `the binary spike at time step $t$ of neuron $j$ in layer $l$'.

The goal of this paper is to reduce $T$ for a rate encoded SNN. To do this, we need to control three types of errors:

\begin{itemize}
    \item {\bf Type 1} errors come from the conversion of ANNs, especially deep ANNs, to SNNs. More precisely, it is the information loss from using discrete spikes instead of continuous FP32-type activation. Note that the issue of quantizing weights is orthogonal and common to both.
    
    \item {\bf Type 2} errors arise from the reduction of time step $T$, especially when $T$ is small. A key insight here is that a typical spiking convolutional layer outputs a spike train in which the spikes are {\em randomly distributed} rather than {\em evenly distributed} in the time window.  An example of a spike train would be [0, 0, 0, 1, 1] for the value of 2/5. As we can see, both spikes are squeezed to the last two time steps. If we now reduce $T$ to say 2, we may lose these last two important spikes.
    
    \item {\bf Type 3} errors are due to the representation skew in the input train spike train when the time steps are extremely small. For example, when the time step is equal to 1, all ANN input $x \in[0,1)$ will be transmitted 0 spikes and only $x=1$ generates a spike. In other words, a zero spike represents a range of values while the 1 spike represents just one specific value, effectively a very skewed quantization, resulting in poor performance.

\end{itemize}

\section{Approach}
\label{sec:approach}

\subsection{Mitigating Type 1 Errors: Value-range (VR) Encoding}
\label{sec:vr encoding}

Type 1 error is the {\em approximation error} caused by the information loss during the ANN-to-SNN conversion because a spike train of size $T$ can only represent a small set of values relative to those represented by 32-bit floating point numbers (FP32) in ANNs. This error is made worse through accumulation in a deep network.  

\begin{figure*}[ht]
    \centering
    \includegraphics[width = 0.85\linewidth]{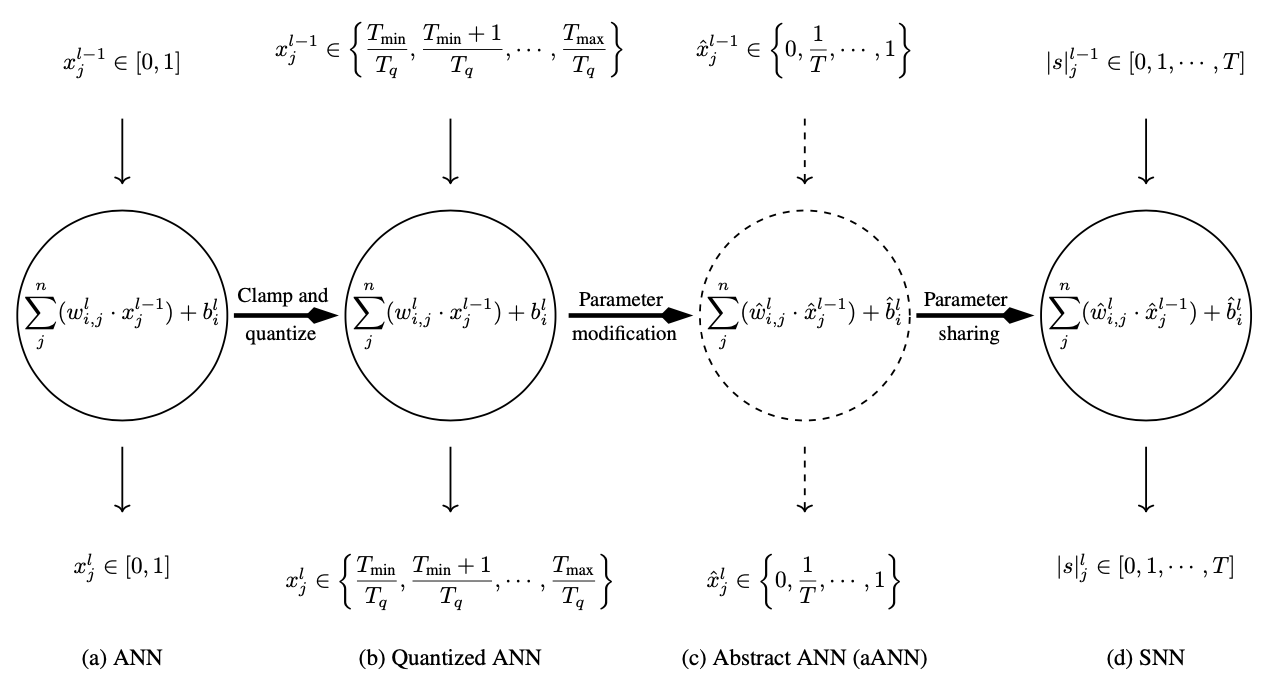}
    \caption{ANN-to-SNN conversion.}
    \label{fig:ann2snn}
\end{figure*}

We shall now describe a novel value-range encoding method to match the values of ANN shown in Fig.~\ref{fig:ann2snn}(b) and its twin SNN shown in Fig.~\ref{fig:ann2snn}(d) exactly to eliminate the approximation error. Fig.~\ref{fig:ann2snn}(a) is the origin ANN while Fig.~\ref{fig:ann2snn}(c) is an abstraction we shall use to show the equivalence. For simplicity, we will avoid writing the superscript and subscript when the context is clear. 

\subsubsection{ANN activation encoding} 
To map an ANN to its twin SNN, we first quantize the activation values (both input and output) of an ANN neuron by down-sampling any FP32 value in $[0, 1]$ to a member of the set $\{0, 1/T_q, 2/T_q,..., T_q / T_q = 1\}$ that is nearest to it, where $T_q$ is the quantization level of ANN and also the desired time window size of the twin SNN. Then, according to insight that values in a ANN can be in an interval smaller than $[0, 1]$, we reduce the time window size further by mapping instead to the set $\{T_{\text {min}}/T_q, T_{\text{min}}+1/T_q,T_{\text{min}}+ 2/T_q,..., T_{\text{max}}/T_q\}$ ($T_{\text{min}}$ and $T_{\text{max}}$ are integers that indicate the 
min and max number in this set). This effectively quantizes the FP32 values to $[T_{\text{min}}/T_q, T_{\text{max}}/T_q]$ instead of $[0,1]$. Any value in $(T_{\text{max}}/T_q, 1]$ is rounded off to $T_{\text{max}}/T_q$, and it is this $T = T_{\text{max}}-T_{\text{min}}$ that will be used as the final time step in SNN inference.

In the ANN training process, we first train the network using ReLU activation for fast convergence. Then, we {\em clamp and quantize} the ANN’s activation values to $\{T_{\text{min}}/T_q, T_{\text{min}}+1/T_q,T_{\text{min}}+ 2/T_q,..., T_{\text{max}}/T_q\}$ and retrain the ANN (details in the Appendix). 


\subsubsection{SNN Weight and Bias Modification}

Next, we need to modify the weights and bias of the SNN (Fig.~\ref{fig:ann2snn}(d)) to map it to its twin ANN (Fig.~\ref{fig:ann2snn}(b)). We shall use an abstract ANN shown in Fig.~\ref{fig:ann2snn}(c) (which we will call `aANN' to distinguish it from Fig.~\ref{fig:ann2snn}(b)) to give insight into the conversion process. The weights and biases of the aANN is the same as the SNN, labeled $\hat{w}, \hat{b}$. 

We shall start by showing the equivalence between the aANN and the SNN. The activation value $\hat{s}_i^l$ (encoded as a spike train with spike rate of $\hat{s}_i^l$) being transmitted from SNN neuron $i$ of layer $l$ is: 
\begin{align}
s^l_i  &= \floor*{\dfrac{1}{\theta^l_i}(\sum_j \hat{w}^l_{i,j} s^{l-1}_j + \dfrac{\hat{b}_i}{T}T)}\\ &= \dfrac{\sum_j \hat{w}^l_{i,j} s^{l-1}_j + \hat{b}_i}{\theta^l_i}+ R^l_i
\end{align}

\noindent
where $R^l_i$ is the residual due to taking the floor and $\theta^l_i$ is the threshold in neuron $i$ of layer $l$.
The aANN quantized the activations of the ANN by steps of $1/T$. This gives us the following: 

\begin{align}
    \hat{x}^{l}_i = \sum_{j=0}^n \hat{w}^l_{i,j} \cdot \hat{x}^{l-1}_{j} + \hat{b}^l_i+ \bar{R}^l_i
\end{align}

\noindent
where $\bar{R}^l_i$ is the quantization error.

If the values of $x$ are in $[0, 1]$, and $\theta^l_i=1$, it is easy to derive that $\bar{R}^l_i\approx R^l_i$ and $s^l_i \approx \hat{x}^{l}_i$, given the 1-to-1 mapping of the values of $|s|$ and $\hat{x}^{l}_i$. The next section will discuss the use of other $\theta$ values.

Then, we can constrain $x$ of the Quantized ANN and $\hat{x}$ of the aANN as:

\begin{equation} 
 \sum_{i=1}^{n} \sum_{k=1}^{k} x_i^k w_i^k + b = \sum_{i=1}^{n} \sum_{k=1}^{k} \hat{x}_i^k \hat{w}_i^k + \hat{b} 
\end{equation}

\noindent
where $k$ are the kernel sizes and $n$ is the number of channels. Since we know $x = \hat{x}(T_{\text{max}}-T_{\text{min}})/T_q $, we have: 
\begin{equation} 
 \sum_{i=1}^{n} \sum_{k=1}^{k} \hat{x}_i^k (\frac{T_{\text{max}}-T_{\text{min}}}{T_q} ) w_i^k+b= \sum_{i=1}^{n} \sum_{k=1}^{k} \hat{x}_i^k \hat{w}_i^k+\hat{b} 
\end{equation}

\noindent
It is now easy to obtain:

\begin{equation} 
\hat{w} = \frac{T_{\text{max}}-T_{\text{min}}}{T_q} w = \frac{T}{T_q} w \;\;\; \text{and}\;\;\; \hat{b} = b +  \sum_{i=1}^{n}\sum_{k=1}^{k}w_i^k \frac{T_{\text{min}}}{T_q} 
\end{equation}

It is these weights and biases, i.e., $\hat{w}$ and $\hat{b}$, that will be used in the twin SNN.

\subsubsection{SNN activation encoding}



Next, we need to map the Quantized ANN's activation value $x \in X = \{T_{\text{min}}/T_q, T_{\text{min}}+1/T_q,T_{\text{min}}+ 2/T_q,..., T_{\text{max}}/T_q\}$ to spike trains that have $|s|$ spikes, $|s| \in S = \{0,1,2,...,T_{\text{max}}- T_{\text{min}}\}$. Note that the cardinality of $X$ is $|s|$ which is exactly $T_{\text{max}}- T_{\text{min}} + 1$. There is therefore a one-to-one mapping between $X$ and $S$. In particular, a value of $j/T_q + T_{\text{min}}/T_q$ would be mapped to $j$ spikes distributed in a spike train in accordance with the IF model. 

\begin{equation} 
\label{v_sT}
x = \dfrac{|s|}{T_q} + \frac{T_{\text{min}}}{T_q}
\end{equation} 


According to Algorithm~\ref{alg1}'s integrate and fire step, the membrane potential is accumulated $T_{\text{max}}-T_{\text{min}}$ times ($T= T_{\text{max}}-T_{\text{min}}$), each time checking against $\theta$ to generate a spike. Using a value from $X$ to approximate the membrane potential, we can get the number of spikes in the SNN as:

\begin{equation} 
\label{svT}
|s| = \frac{x(T_{\text{max}}-T_{\text{min}})}{\theta} 
\end{equation}

\noindent
By combining Equation~\ref{v_sT} and Equation~\ref{svT}, we get:

\begin{align} 
\theta = \frac{x (T_{\text{max}}-T_{\text{min}})}{|s|} = \frac{(\dfrac{|s|}{T_q}+\dfrac{T_{\text{min}}}{T_q}) (T_{\text{max}}-T_{\text{min}})}{|s|} 
\end{align}

\noindent
Then, we subtract $\dfrac{T_{\text{min}}}{T_q}$ from $x$  at each time step and get: 

\begin{align} 
\label{threshold}
\theta =  \frac{(\dfrac{|s|}{T_q}+\dfrac{T_{\text{min}}}{T_q}-\dfrac{T_{\text{min}}}{T_q}) (T_{\text{max}}-T_{\text{min}})}{|s|} = \dfrac{T_{\text{max}}-T_{\text{min}}}{T_q}
\end{align}
\noindent
In other words, if we use the $\theta$ given in Equation~\ref{threshold} in Step 2 of Algorithm~\ref{alg1}, we will generate a spike train of $|s|$ evenly distributed spikes. To keep things simple, we have ignored any error in the use of a value of $X$ to approximate the FP32 membrane potential in the derivation of $\theta$. Our experiments showed that this assumption of negligible impact is valid.

\subsection{Mitigating Type 2 Errors: Averaging IF Spike Generation (ASG) Model}
\label{sec:ASG model}

In this section, we will focus on solving the problem of Type 2 error that arises from reducing $T$. To do this, we introduce a novel spike encoding method we call the {\em averaging IF spike generation} (ASG) model that introduces an averaging step not found in the standard IF model. In particular, in the ASG model, a neuron, $i$ say, in layer $l$ first computes the {\em membrane potential} $V^l_i(t)$, at time step $t$, just as in the standard IF model. This is the weighted sum of its corresponding $j$ weights $w^l_{i,j}$ of layer $l$, and the input spikes $s_j^{l-1}(t)$ received at each time step $t$. A bias $b^l_i$ is added. 
Then, we sum up the membrane potential $V^l_i(t)$ at each time step and divide the sum by the time window size $T$ to get an average membrane potential $A_i^l$ which is the same at each time step (the division operations can be merged into weights and biases of the previous layer since it's a linear operation). Finally, we send $A_i^l$ to the IF model, at each time step $t$, a neuron $i$  `integrates' (sums)  $A_i^l$ of the previous time steps as the membrane potential $U_i^l$. Whenever $U_i^l$ exceeds a predefined threshold $\theta$, the neuron fires a spike of `$1$', and decreases $V_i^l$ by $\theta$. Otherwise, it outputs `$0$'.

\begin{algorithm}[ht]
\caption{Averaging IF spike generation model.} 
\label{alg1}
\begin{algorithmic}[1]
\REQUIRE Weights $\bm{w^l_{i,j}}$ between neuron $i$ at layer $l$ and neuron $j$ at layer $l-1$; Input spikes $s_j^{l}$ at neuron $j$ at layer $l$; Bias $b^l_i$ at neuron $i$ at layer $l$; Membrane potential $U_i^l$ of neuron $i$ at layer $l$ (initialize to 0). Default threshold $\theta$.
\STATE \textbf{Step 1:} Averaging Operation:
\FOR{$t=1$ to $T$}
\STATE $V_i^l(t) = \sum_{j} (w^l_{i,j} \cdot s_j^{l-1}(t) + b^l_i)$
\ENDFOR
\STATE $A_i^l = (\sum_{t=1}^{T}V_i^l(t))/T $

~

\STATE \textbf{Step 2:} Integrate and fire:
\STATE $s_i^l(0) = U_i^l(0) = 0$;
\FOR{$t=1$ to $T$}
\STATE $U_i^l(t) = U_i^l(t-1) +A_i^l;$
\STATE $
s_i^l(t)=\begin{cases}
1, & U_i^l(t) \geq \theta\\
0, & \text{otherwise}
\end{cases};
$
\STATE $U_i^l(t)=\begin{cases}
U_i^l(t) - \theta, & U_i^l(t) > \theta\\
U_i^l(t), & \text{otherwise}
\end{cases}.$ 
\ENDFOR
\STATE \textbf{return} $s_i^l$ - the ASG spike train of neuron $i$ of layer $l$. 

\end{algorithmic}
\end{algorithm}

With ASG, each spiking convolutional layer outputs a spike train in which spikes are evenly distributed rather than randomly distributed in the time window. This improves the robustness of low latency SNNs. For example, one SNN neuron with a time step of 5 has the following membrane potential in each time step: [0.1, 0.2, 0.5, 0.8, 0.9]. In the traditional IF model, the spike train will be [0, 0, 0, 1, 1] when the threshold is 1. As we can see in this example, both spikes are squeezed to the last two time steps. If we now reduce $T$ to say 2, we will lose these last two spikes, hence impacting accuracy. Also, such Type 2 error will accumulate over the layers and causes a large loss in accuracy.


The key contribution of ASG is to reshape the spike distribution to avoid this issue. For the same example, after the averaging operation, the membrane potential will be [0.5, 0.5, 0.5, 0.5, 0.5], leading to a spike train of [0, 1, 0, 1, 0] after IF model. Such an evenly distributed spike train is more resilient to reductions of $T$. 

\subsection{SNN Threshold Training Method}
\label{sec:SNN tt}

By combining VR encoding with ASG model, we can eliminate both types 1 and 2 errors, and achieve near-zero accuracy drop after conversion with a significantly reduced $T$. However, the ASG model requires all membrane potential at different time steps to be known so as to compute the average. To solve this problem, we shall now introduce a channel-wise {\em threshold training} (CTT) method that uses unique thresholds for each channel. It essentially performs tandem training of an equivalent SNN that uses the traditional IF model alongside its twin ASG model. The main goal is to learn spike firing thresholds that best narrow the gap between the two models. Instead of a constant firing threshold, unique thresholds are derived for each channel in the IF model SNN.

As shown in Algorithm~\ref{alg_tt}, we define a loss function $L$ which is the difference in the number of spikes between the output spike train of the IF model $s_1^{l_i}$ and the ASG model $s_2^{l_i}$. This loss function incentivizes the closing of the gap between the number of spikes in the two spike trains. 

\begin{algorithm}[t]

    \centering
    \caption{Threshold training for IF SNN:}
    \label{alg_tt}
\begin{algorithmic}[1]
\REQUIRE Input $\bm{x}$ of a spike layer ;
\ENSURE $N$ spike layers; $i_{th}$ layer has $k_i$ channels. threshold of ASG model $T/T_q$
\FOR{$i=1$ to $N$}
\FOR{$k=1$ to $k_i$}
\STATE $\bm{s}_1^{l_i^k} = \textbf{IF Model}(\bm{x}^{l_i^k},\bm{\theta^{l_i^k}})$
\STATE $\bm{s}_2^{l_i^k} = \textbf{ASG Model}(\bm{x}^{l_i^k},\bm{\theta}= T/T_q)$
\STATE  $L = \sum s_2^{l_i^k} - \sum s_1^{l_i^k}$ // Difference of \#spikes
\STATE $\theta^{l_i^k} = \theta^{l_i^k} - lr*L $ 
\ENDFOR
\ENDFOR

\end{algorithmic}
\end{algorithm}

\subsection{Mitigating Type 3 Errors: Input Channel Expansion (ICE)}
\label{sec:ice}

To mitigate type 3 errors, we propose {\em input channel expansion} (ICE) that improves the robustness of the input data. As described in Algorithm~\ref{alg_ice}, the normalized input $x$ is first quantized to $x_q\in \{T_{\text{min}}/T_q, T_{\text{min}}+1/T_q,T_{\text{min}}+ 2/T_q,..., T_{\text{max}}/T_q\}$. When $T = T_{\text{max}}-T_{\text{min}}$ has been reduced to a small number, for example, 1, such a quantized $x_q$ may not be enough to represent the input due to the information loss during quantization. Hence, we keep augmenting the input data by quantizing it at different levels, for example, $T_q+1$ and we can get $x'_q \in \{T_{\text{min}}/(T_q+1), T_{\text{min}}+1/(T_q+1),T_{\text{min}}+ 2/(T_q+1),..., T_{\text{max}}/(T_q+1)\}$. To maintain the consistency of the spike train of SNNs converted from the quantized input data, we turn values in $x'_q$ expressed as $t/(T_q+1)$ into $t/(T_q)$, shown in Algorithm~\ref{alg_ice}. After $\phi$ times of concatenation, we get the final input $x_q$ which has $\phi$ times more channels compared with the original input. Finally, for the SNN, we repeatedly feed the concatenated $x_q$ into a spike layer of threshold $(T_{\text{max}} - T_{\text{min}})/T_q$ for $T$ time steps to obtain the input spike train.

\begin{algorithm}[t]
    \centering
    \caption{Input channel expansion (ICE).}\label{algorithm1}
    \label{alg_ice}
    \begin{algorithmic}[1]
\REQUIRE Input data $x \in [0,1]$. \\
\qquad\enspace Input channel expansion factor $\phi$.  \\
\ENSURE Spike train $\bm{s}$ of length $T$. Output $x_o$.\\

\STATE $x_o = [ \; ]$ 
\FOR{ $c= 1$ to $\phi$}
\STATE $x_q = \lfloor x\cdot T\rfloor/T$ 
\STATE $x'_q = \lfloor x\cdot (T+1)\rfloor/(T+1)$
\STATE $\hat{x_q} = \lfloor x\cdot (T-1)\rfloor/(T-1)$
\FOR{ $t= 1$ to $T$}
\STATE $where(x'_q == t/(T+1)),t/(T)$ 

\STATE $where(\hat{x}_q == t/(T-1)),t/(T)$ 
\ENDFOR
\STATE $x_o = Concat(x_o,x_q,x'_q,\hat{x}_q)$ 
\ENDFOR

\end{algorithmic}
\end{algorithm}

\section{Experiment}
\label{sec:exp}
\subsection{Experiment Setup}

We have implemented our schemes in CUDA-accelerated (CUDA 11.7) PyTorch version 1.13.0. The experiments were performed on an Intel Xeon E5-2680 server with a Tesla P100 GPU and a GeForce RT 3090 GPU, running 64-bit Linux 5.15. We adopt LeNet-* (a 7-layers Lenet type network) on MNIST dataset\footnote{http://yann.lecun.com/exdb/mnist/}, VGG-*~\cite{yan2021near} (a 7-layers plain VGG) and VGG-16 on CIFAR10/100\footnote{https://www.cs.toronto.edu/~kriz/cifar.html}, and RepVGG-B3 (a 28-layers plain VGG) ~\cite{ding2021repvgg} and VGG-16 on ImageNet\footnote{https://image-net.org/challenges/LSVRC/2012/index.php}.
\subsection{Experiment Results}

\begin{table*}[hbt]
\centering
\scalebox{0.85}{
\begin{tabular}{cccccc}
\hline
Dataset                        & Methods                                                                                        & SNN Accuracy & Architecture                                               & Time Steps &$\varepsilon_{\text{SNN}}$ \\ \hline

\multicolumn{1}{c|}{} &ANN-to-SNN~\cite{bu2022optimized}                                                               & \underline{90.96}/94.20\%          & \underline{VGG-16}                                                   & \underline{8}/32    &3.61/0.37\%   \\
\multicolumn{1}{c|}{}          & ANN-to-SNN ~\cite{deng2021optimal}            & 92.29/92.24\%          & VGG-16                                                       & 16/128 & -0.2/-0.15\%       \\

\multicolumn{1}{c|}{}          &ANN-to-SNN 2021~\cite{li2021free}         & 93.71/95.65\%             & VGG-16  &32/128 &2.01/0.07\% \\

\multicolumn{1}{c|}{}          &ANN-to-SNN~\cite{yan2021near}   \multirow{4}{*}{}          & 94.16\%       & VGG-$*$    & 600  &0.04\%   \\
\multicolumn{1}{c|}{CIFAR-10}          & ANN-to-SNN~\cite{yu2021constructing}            & 92.76\%          & VGG-$*$    & 500    & 0.66\% \\
\multicolumn{1}{c|}{}          & Tandem Learning\cite{wu2021tandem}            & 90.98\%          & Cifar-Net    & 8    & 0.79\% \\
\multicolumn{1}{c|}{}          & ANN-to-SNN\cite{han2020rmp}            & 93.63\%          & VGG-16   & 2048 &0.01 \%    \\
\multicolumn{1}{c|}{}          & Hybrid Training\cite{rathi2020enabling}            & 92.02\%          & VGG-16   & 200 &/    \\
\multicolumn{1}{c|}{}          & Spike-based BP\cite{wu2019direct}            & 90.53\%          & 5 Conv    & 12 &/    \\

\cline{2-6}

\multicolumn{1}{c|}{}          &                             & \textbf{84.14/93.71\%}   & VGG-$*$                                                    & \textbf{1/8}   &-0.11/0\%   \\

\multicolumn{1}{c|}{}          & This paper (VR+ICE+ASG)                        & \textbf{94.67/94.81\%}   & VGG-$*$                                                    & \textbf{16/32}     &-0.03/-0.01\%   \\
\multicolumn{1}{c|}{}          &                       & \underline{\textbf{91.55}}\textbf{/92.33\%}   & \underline{VGG-16}                                                    & \underline{\textbf{8}}\textbf{/16}     &0.01/0.05\%   \\
\cline{2-6}
\multicolumn{1}{c|}{}          & This paper (VR+ICE+CTT)                            & \textbf{86.04/91.43\%}   & VGG-$*$                                                    & \textbf{16/32}   &8.92/3.37\%   \\

 \hline\hline
 
\multicolumn{1}{c|}{} &ANN-to-SNN~\cite{bu2022optimized}                                                               & \underline{60.49}/74.82\%          & \underline{VGG-16}                                                   & \underline{8}/32    &15.82/1.49\%  \\
\multicolumn{1}{c|}{}          & ANN-to-SNN ~\cite{deng2021optimal}                                                                            &65.94/70.47\% & VGG-16                                                      & 16/128 & 4.68/0.15\%     \\
\multicolumn{1}{c|}{}          & ANN-to-SNN~\cite{yan2021near}                                                                           & 71.84\% & VGG-$*$                                                       & 300    & 0\%  \\
\multicolumn{1}{c|}{CIFAR-100}          & ANN-to-SNN~\cite{li2021free}             & 73.55/77.40\%          & VGG-16    & 32/128 &4.34/0.49\%   \\

\multicolumn{1}{c|}{}          & ANN-to-SNN~\cite{han2020rmp}                                                                     & 70.93\% & VGG-16                                                      & 2048 &0.29\%    \\

\multicolumn{1}{c|}{}          & ANN-to-SNN~\cite{sengupta2019going}                                                                     & 65.47\% & ResNet-34                                                   & 2000 &6.12\%    \\

\multicolumn{1}{c|}{}          & Spike-based BP\cite{li2021differentiable}                                                                     & 74.24\% & ResNet-18                                                      & 6 &/\    \\

\cline{2-6}

\multicolumn{1}{c|}{}          &                                                                           & \textbf{54.49/76.38\%} & VGG-$*$                                                      & \textbf{1/8}   & 0.12/-0.16\%  \\
\multicolumn{1}{c|}{}          &     This paper (VR+ICE+ASG)                  & \textbf{77.36\%} & VGG-$*$                                                      &\textbf{16}   &0.14\%    \\
\multicolumn{1}{c|}{}          &                      & \underline{\textbf{64.79}}\textbf{/66.11\%} & \underline{VGG-16}                                                      &\underline{\textbf{8}}\textbf{/16}   &-0.02/0.15\%    \\\cline{2-6}
\multicolumn{1}{c|}{}          & This paper (VR+ICE+CTT)                            & \textbf{66.34/72.60\%}   & VGG-$*$                                                    & \textbf{16/32}   &9.67/4.72\%   \\

\hline\hline

\multicolumn{1}{c|}{} &ANN-to-SNN~\cite{bu2022optimized}           & \multirow{4}{*}{}                                                        74.24/74.62\%          & VGG-16                                                   & 128/256  &0.61/0.23\%   \\
\multicolumn{1}{c|}{}          &  ANN-to-SNN ~\cite{deng2021optimal}                                                                          &\underline{67.73}/72.34\% & \underline{VGG-16}                                                      & \underline{32}/512   & 4.67/0.06\%  \\

\multicolumn{1}{c|}{ImageNet}          & ANN-to-SNN~\cite{li2021free}                                                                        & 63.64/74.23\% & VGG-16                                                       & 32/256 & 11.72/1.13\%     \\
\multicolumn{1}{c|}{}          & ANN-to-SNN\cite{han2020rmp}           & 73.09\%          & VGG-16    & 4096  &0.4\%  \\
\multicolumn{1}{c|}{}          & Spike-based BP\cite{fang2021deep}           & 68.76\%          & SEW ResNet-101   & 4  &/\   \\
\multicolumn{1}{c|}{}          & STBP-tdBN\cite{zheng2021going}           & 63.72\%          & ResNet-34   & 6  &/\    \\

\cline{2-6}

\multicolumn{1}{c|}{}          &      This paper (VR+ICE+ASG)                                                                     &  \textbf{75.35/79.16\%} & Rep-VGGB3                                                      & \textbf{100/200}   &0.09/0\%   \\
\multicolumn{1}{c|}{}          &        This paper (VR+ICE+ASG)           & \underline{\textbf{70.43\%}} & \underline{VGG-16}                                                      &\underline{\textbf{32}}   &1.12\%   \\\cline{2-6}
\multicolumn{1}{c|}{}          &        This paper (VR+ICE+CTT)           & \textbf{68.16\%} & VGG-16                                                      &\textbf{160}   &3.38\%   \\
\hline

\end{tabular}}

\caption{Comparison with state-of-the-art transfer learning on SNNs CIFAR-10, CIFAR100, and ImageNet dataset. In all cases, all parameters are in FP32. Compatible comparisons are underlined.}
\label{comparsion on cifar10}
\end{table*}
 Our full set of experiment results is shown in Table~\ref{comparsion on cifar10}. We define $\varepsilon_{\text{SNN}}$ to be the accuracy difference between the VR-encoded ANN (Fig.~\ref{fig:ann2snn}(b)) and the final SNN.

In this paper, SNN accuracy of 93.71\% with $T = 8$ with $\varepsilon_{\text{SNN}} = 0$\% was achieved for CIFAR-10. We can also achieve an SNN accuracy of 94.67\% if $T=16$ ($\varepsilon_{\text{SNN}} = -0.03$\%). For CIFAR-100, our SNN achieved 76.38\% accuracy with $T = 8$ and $\varepsilon_{\text{SNN}} =  -0.17$\%. At an extremely small $T=1$, we can also achieve comparable SNN accuracy with almost no accuracy drop: -0.09\% on CIFAR-10 and 0.13\% on the CIFAR-100 dataset. 
With a Lenet-like network structure,  running on the MNIST dataset, we can achieve an SNN accuracy of 98.73\% with $T=1$. We also applied our techniques to a much deeper network structure with a larger dataset to test whether they work on complex models. For the ImageNet dataset, an SNN accuracy of 75.35\% with $T = 100$ was achieved with $\varepsilon_{\text{SNN}} = 0.09$\%. At $T= 32$, we can still achieve an SNN accuracy of 70.43\% with an accuracy drop of 1.12\%. 

Using our CTT training method to obtain standard IF model SNNs, we achieved SNN accuracies of 86.04/91.43\% (CIFAR-10) and 66.34/72.60\% (CIFAR-100) with slightly larger $T$ of 16 and 32, respectively. These accuracies are still higher than most existing works and achieved using much smaller time window sizes. Running on ImageNet, SNN accuracy of 68.16\% with an ANN-to-SNN accuracy drop of 3.38\% was achieved at $T=160$.

\subsection{Comparison with the state-of-the-art}

Deng and Gu~\cite{deng2021optimal} used threshold ReLU and threshold shift to achieve an SNN accuracy of 92.24\% and 70.47\% for the CIFAR-10 and the CIFAR-100 datasets, respectively, with $T = 128$. Han {\em et al.}~\cite{han2020rmp} proposed a residual membrane potential (RMP) spiking neuron model, and an SNN accuracy of 93.63\% and 70.93\% was achieved on the CIFAR-10 and CIFAR-10 dataset, respectively. $T$ used here was 2048. Yu {\em et al.}~\cite{yu2021constructing} used a typical double threshold balancing, Li {\em et al.}~\cite{li2021free} proposed  layer-by-layer calibration algorithm and Yan {\em et al.}~\cite{yan2021near} used clamp and quantization ANN pre-training. All these three works can achieve high SNN accuracy but with a large time step. For example, $T$ was 600 for CIFAR-10 and 300 for CIFAR-100 in Yan {\em et al.}~\cite{yan2021near}.  For directly training, running on the ImageNet dataset, Zheng{\em et al.}~\cite{zheng2021going}, Fang{\em et al.}~\cite{fang2021deep} and meng{\em et al.}~\cite{meng2022training} achieved SNN accuracies of 63.72\%, 68.76\% and 67.74\% based on STBP-tdBN, Spike-based BP and differentiation on spike representation, respectively, which are always lower than ANN-to-SNN methods ~\cite{deng2021optimal,han2020rmp} but with smaller time steps. 

For variants of the standard SNNs, Bu {\em et al.}~\cite{bu2022optimized,bu2021optimal} skipped SNN input encoding and used constant float32-type input of the test images (which means global information is known at the very first time step and additional multiplication operations are needed in the first convolutional layer). SNN accuracies of 90.96\% on CIFAR-10 ($T=8$), 60.49\% on CIFAR-100 dataset ($T=8$) were achieved. Running on ImageNet, SNN accuracies of 64.70\% and 74.24\% were achieved with time steps of 32 and 128. For direct training (Spike-based BP), Rathi~\cite{rathi2021diet} and Li~\cite{li2021differentiable} achieved SNN accuracies of 69.00\% and 71.24\% ($T=5$) but with softmax in the last layer which may lead to additional multiplication operations. In this paper, after mitigating all type errors, we can achieve higher SNN accuracies with smaller time steps, a smaller VGG network structure -- the number of parameters of the VGG-* is $0.68\times$ that of VGG-16, and no more additional multiplication operations when compared with other state-of-the-art works on both CIFAR-10/100 and Imagenet datasets. Additionally, for ImageNet, we can achieve a much higher SNN accuracy up to 79.16\% with $\varepsilon_{\text{SNN}} = 0.09$\% using a deep Rep-VGGb3 model. 

In the extreme, {\em binary neural networks} (BNNs) replace floating-point multiplication and addition operations with XNOR-bitcount operations. Unfortunately, as shown in the survey~\cite{qin2020binary}, running on the CIFAR-10 dataset, BNN accuracies vary from 66.6\% to 88.6\%, which is lower than most SNNs. Also, BNNs have float32-type inputs, hence we did not consider it for comparison.

\begin{figure*}[t] 
\begin{minipage}[t]{0.2455\linewidth} 
\centering
\includegraphics[width = 1.0\linewidth]{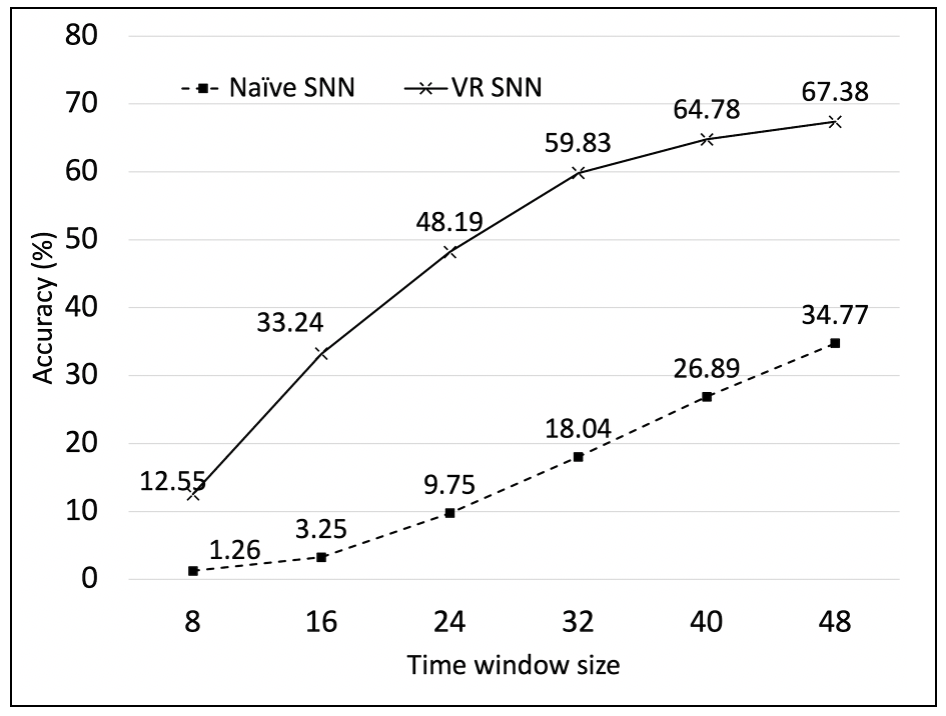}
\caption{VR ablation study} 
\label{fig_vr_} 
\end{minipage}
\begin{minipage}[t]{0.245\linewidth}
\centering
\includegraphics[width = 1.0\linewidth]{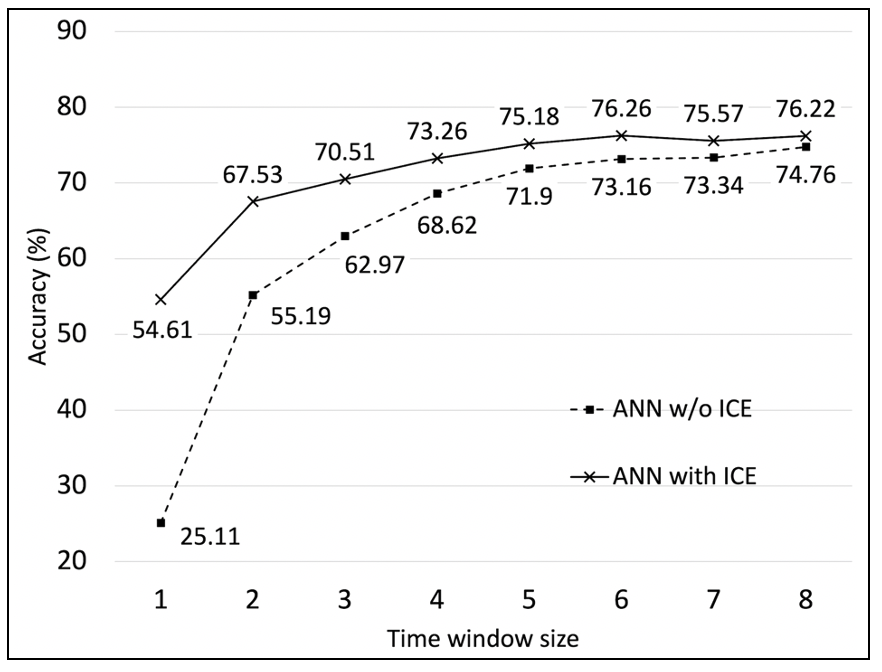}
\caption{ICE ablation study}
\label{fig_ice}
\end{minipage}
\begin{minipage}[t]{0.2455\linewidth}
\centering
\includegraphics[width = 1.0\linewidth]{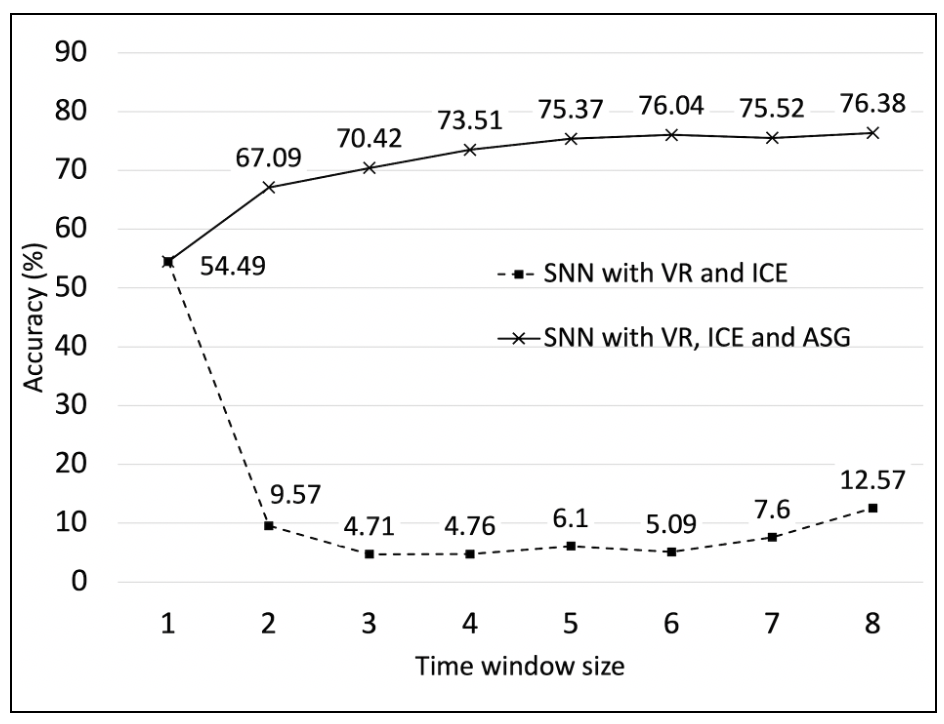}
\caption{ASG ablation study}
\label{fig_asg}
\end{minipage}
\begin{minipage}[t]{0.2455\linewidth}
\centering
\includegraphics[width = 1.0\linewidth]{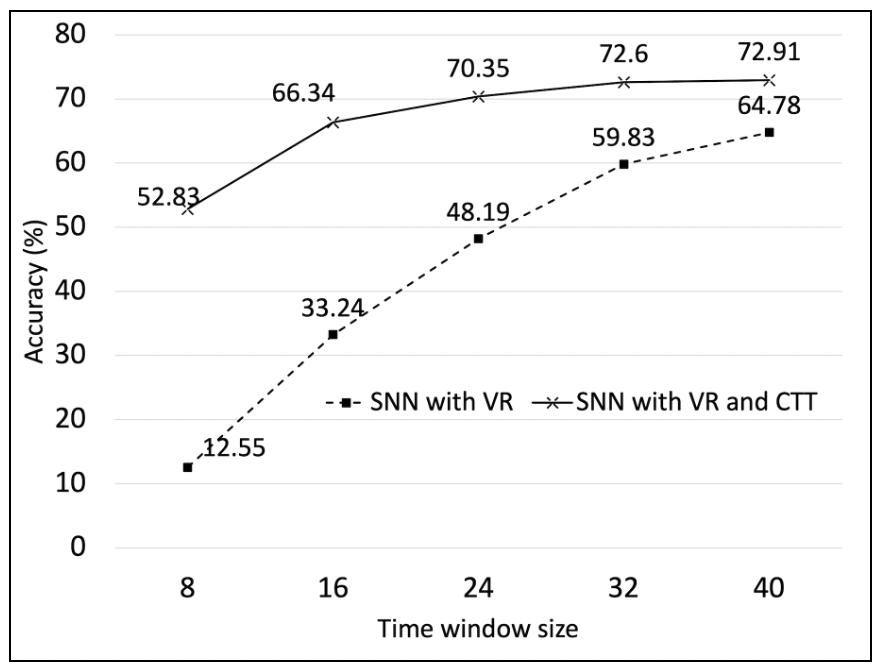}
\caption{CTT ablation study}
\label{fig_ctt}
\end{minipage}
\end{figure*}

\subsection{Full Ablation Study}
For an ablation study, we first transferred the weights of the VGG-$*$ (with a baseline full precision accuracy of 77.59\%) trained for the CIFAR-100 dataset na\"{\i}vely to a twin SNN directly. This is compared with an SNN converted via VR-encoded ANN in Figure ~\ref{fig_vr_}. As most of values in a ANN are 0, $T_{\text{min}}/T_q$ was set to 0. $T_{\text{max}}/T_q$ is set as 0.8. This is the smallest scaling factor we found that does not impact the baseline ANN accuracy after retraining. Testing on VGG-$*$ on the CIFAR-100 dataset, the accuracy difference before and after retraining can be enlarged from 0.05\% to 3.48\% if we further reduced $T_2/T_1$ from 0.8 to 0.7. As we could see in Fig.~\ref{fig_vr_}, SNN accuracies with VR-encoding are roughly 11.29\% to 41.79\% higher than na\"{\i}ve transfer. Next, ICE was introduced into the VR-encoded ANN, and $T$ is reduced. We set the ICE expansion factor $\phi$ to be 2 because we found that larger values were not useful. The number of parameters in the VR-encoded ANN, ICE-based VGG-$*$ is only $1.0018\times$ that of the original VGG-*. As shown in Figure ~\ref{fig_ice}, ANN accuracies with ICE are 29.5\% higher than without ICE at $T=1$. The accuracy gap drops when $T$ is increased. We then converted the ANN model to SNN using the ASG model. As shown in Figs.~\ref{fig_ice} and~\ref{fig_asg}, the accuracy loss when the VR-encoded, ICE-enhanced ANN is converted to SNN using ASG, $\varepsilon_{\text{SNN}}$, is practically zero. The final accuracies are comparable to the state-of-art SNN ones. For example, we achieve an SNN accuracy of 76.38\% with a latency of 8 and $\varepsilon_{\text{SNN}} = -0.16$\%. Even for $T = 1$, we can achieve an SNN accuracy of 54.49\% with $\varepsilon_{\text{SNN}} = 0.12$\%. Finally, CTT was used to tandem-train an SNN that uses the traditional IF model alongside its twin ASG model. As shown in Figure ~\ref{fig_ctt}, SNN accuracies with CTT are 40.28\% higher than without CTT when the time step is equal to 8. We can also achieve an SNN accuracy of 72.6\% with the standard IF model at $T = 32$.

\section{Discussion}
\label{sec:discuss}
\subsection{Weight Quantization}
\label{sec:quantization}

Thus far, we have only quantized activations while the weights are in FP32 precision. We have also studied the impact of quantizing the weights of the SNN to fixed point values~\cite{yan2021near}. For 7-bit VGG-$*$ SNNs, running on CIFAR-10 and CIFAR-100 dataset, we achieved accuracies of 93.55\% ($T = 8$) and 75.84\% ($T = 8$), respectively. $\varepsilon_{\text{SNN}}$ for these datasets are near zero. However, it increases to over 1\% if the weights are less than 6-bit.

\subsection{Energy Savings}
We did not implement the models we obtained on any SNN hardware and hence cannot provide absolute energy numbers. Instead, we can get an indication of energy savings based on the average number of `$1$' spikes in the entire time window since `$0$' spikes result in no computation and hence cost no (dynamic) power. With $T=8$, our VGG-$*$ model resulted in an average of
 $476,035$ `$1$' spikes out of a total possible $4,169,728$ (i.e. $11.4$\%) spikes for a single test image of CIFAR-100. On the other hand, using the same model from Yan {\em et al.}~\cite{yan2021near} that has $T = 300$ resulted in $17,965,810$ out of $117,043,200$ (or $15.3$\%) that are `$1$'s. In summary, for the same model, the average number of `$1$' spikes when $T=8$ is a mere $2.65$\% of when $T=300$. Furthermore, the former is 4.54\% higher in accuracy. If we reduce $T$ further from $8$ to $7$, we found that the average number of `$1$' spikes is reduced by 8.57\%. This shows the importance of reducing $T$. It is also noteworthy that the ANN version of the VGG-$*$ model had, on average, 28.02\% non-zero activation values when running CIFAR-100, each of which requires a full multiplication. 
 

\subsection{Verification on Other Network Structures}

Jeffares{\em et al.}~\cite{jeffares2021spike} showed that SNNs can also be applied to conventional ANNs such as LSTMs, and leads to computational savings on the MNIST dataset. We also implemented our techniques as a plug-and-play ANN-to-SNN library that works for most of the state-of-the-art neural network structures. 
On the same MNIST dataset, SNNs created from ResNet, MobileNet and DenseNet blocks of our library achieved SNN accuracies of 96.12\%, 95.65\%, and 97.52\%, respectively, with $T=10$. The network structures are shown in the Appendix.

\section{Conclusion}
\label{sec:conclusion}
In this paper, we consider the problem of achieving low latencies (or equivalently, small time window sizes) when converting ANNs to rate-encoded SNNs equivalents. We identified the three main sources of accuracy loss in such conversions, and proposed novel techniques, namely value-range (VR) encoding, averaging IF spike generation (ASG), and input channel expansion (ICE), to mitigate each. We further rounded up our suite of techniques with an SNN threshold training method to derive standard IF model SNN models. Together, they can reduce time window sizes from thousands by nearly two orders of magnitude while maintaining state-of-art accuracies. Our techniques work for the most popular ANN network structures, allowing users to first develop ANN models using existing ANN ecosystems, and train them efficiently before converting them to more energy-efficient SNNs for deployment. All our results are reproducible using the code we have made public.

\bibliography{aaai23}
\subsection{Appendices.}
\subsection{Details of the ANN Training Process}

In this paper, based on VR-encoding, we start with a novel training method that improves the baseline ANN accuracy as followed:

\begin{enumerate}
    \item {\em Pre-training for initial weights}: Having a set of good initial weights significantly impacts the final CNN results. So, for those small datasets like the CIFAR-10 dataset and CIFAR-100 dataset, we first build a bigger model that includes the model for small datasets. The bigger model is trained on the whole ImageNet dataset and has three more convolutional layers to perform downsampling. After this training is completed, the corresponding weights are shared with the model for the smaller dataset, as the initial weights.
    
     
    
    \item 
    {\em Dataset augmentation}: For small datasets like CIFAR-10 and CIFAR-100 datasets, we perform data augmentation to prevent under-training. The new larger training datasets are constructed using the following steps. First, the image is rotated by a random angle ranging from $-15^{\circ}$ to $15^{\circ}$. Then, the origin pictures are enlarged by four to six pixels on each size and randomly cropped to a size expected of the dataset, say, $32\times 32$. After the preprocessing, the modified image is then randomly horizontally flipped with a probability of $0.5$. Gaussian noise with a variance of 0.0001 and a mean of 0 is added to the picture $x$ as the last step, shown in Equation~\ref{eq:guassian}. 
    
    \begin{equation}
    \label{eq:guassian}
    x = \sigma \frac{1}{\sqrt{2\pi}}e^{-\frac{x^2}{2}} + \mu
    \end{equation}
    
    \noindent
    *$\sigma$ and  $\mu$ indicate the variance and mean respectively. 
    
    By applying the above steps repeatedly to the original images in the dataset, we can produce an augmented dataset that is 50 to 100 times larger.
    
    \item To prevent over-fitting, an ANN is trained with both batch normalization and dropout layers. We set the dropout rate as a large number, for example, $0.2-0.4$. RELU activation and average pooling layer are used in this step. 
    
    \item Next, we reduce the dropout rate to a smaller number for higher testing data accuracy. The RELU activation function is replaced by clamping, $C$, (Equation~\ref{eq:clamp}) and quantization, $Q$, (Equation~\ref{eq:quantization}) function sequentially as explained in Section 3.1. After retraining, we have an ANN that is compatible with its twin SNN.
    
    \begin{equation}
    \label{eq:clamp}
    C(x) = \begin{cases}
    \dfrac{T_{\text{min}}}{T_q}, & 0 \leq x \leq \dfrac{T_{\text{min}}}{T_q}\\
    x, & \dfrac{T_{\text{min}}}{T_q} \leq x \leq \dfrac{T_{\text{max}}}{T_q}\\
    \dfrac{T_{\text{max}}}{T_q},& x>\dfrac{T_{\text{max}}}{T_q}. \\
    \end{cases}
    \end{equation}
    
    The gradient of the clamp function is similar to the RELU function. If $x$ is between 0 to 1, the gradient is set as 1, otherwise 0.
    Quantization to a length $T_q$ is performed by 
    \begin{equation}
    \label{eq:quantization}
        Q(x) = \lfloor x\cdot T_q\rfloor/T_q.
    \end{equation}
        
    \item Finally, batch normalization is merged into the ANN trained weights and bias before they are transferred to the SNN as shown in Equation ~\ref{eq:bn_1} and Equation ~\ref{eq:bn_2}.
    
     \begin{equation}
     \label{eq:bn_1}
    \bm{W}^l = \frac{\gamma^l}{\sqrt{(\sigma^l)^2 + \epsilon}}\bm{W}^l 
    \end{equation}
    
     \begin{equation}
      \label{eq:bn_2}
        \bm{b}^l = \frac{\gamma^l}{\sqrt{(\sigma^l)^2 + \epsilon}}(\bm{b}^l - \mu^l) + \beta^l
        \end{equation}
     * $\mu$ and $\sigma$ indicate the mean and variance of $\bm{x}$ in one batch size which learned during ANN training; $\gamma$ and $\beta$ are training parameters stored through training; $x_i^l$ means the parameter $x$ is at convolution layer $l$. $\epsilon$ is set as a fixed number: $10^{-5}$.
     
\end{enumerate}
The resultant SNN can be used for inference as it is. Also, it can be served as the starting point for our next step, i.e., threshold training.

We will take the training process using VGG-$*$ on the CIFAR-100 dataset as an example to show the power of our training method: naively using VR-encoding will lead to an SNN accuracy of only 58.09\% with 30 epochs of training. By combining VR-encoding with data augmentation (Step 2), the testing accuracy of SNN increases to 72.22\%. By adding pre-training for initial weights, we achieved a further 5.02\% increase in accuracy to 77.27\%. 

\subsection{Network Models}

We experimented with some well-known ANN structures including VGG, ResNet, MobileNet v2 and DenseNet. All max-pooling layers in these structures are replaced with average-pooling layers as the rate-encoded SNN does not support max-pooling. Full 32-bit floating point precision was used. 

\begin{table}[ht]
\small
\centering
\begin{threeparttable}
\begin{tabular}{c|l}
\hline
ResNet Block & 32, 32, 32, $R$, $A$  \\\hline
MobileNetV2 Block   & 32, 64, $P$, 64, $D$, 32, $P$, $R$, $A$, $FC$\\ \hline
DenseNet    Block     &  32, 32, $\text{\em Cat}$, 64, $\text{\em Cat}$, 128, $R$, $A$, $FC$ \\ \hline
LeNet*    &32, 32, $C$, 64, 64, $C$, 128\\\hline
VGG-16    &  2$\times$64, $A$, 2$\times$128, $A$, 3$\times$256, $A$,  3$\times$512, \\ & $A$,  3$\times$512, $A$\\\hline
VGG-$*$ & 2$\times$ 128, $A$, 2$\times$256, $A$, 2$\times$512, $A$,\\ & 1024, $A$\\\hline
RepVGG-B3 & 64,4$\times$192,6$\times$384,16$\times$768,1$\times$2560 $A$\\\hline

\end{tabular}
\caption{Summary of network structures.}
\label{tab:network}
\end{threeparttable}
\end{table}

The network structures and blocks we used are summarized in Table~\ref{tab:network}. The last dense classifier layer, present in all the networks, is omitted; `$R$' stands for a connection from previous layers; `$R_C$' stands for a shortcut from previous layers; `$P$' stands for pointwise convolution; `$D$' stands for depthwise convolution; `$\text{\em Cat}$' stands for a concatenation from previous layers. `$A$' stands for an average pooling layer. `$C$' stands for a stride-2 convolutional layer used for down-sampling.`$FC$' stands for a full connection layer.

\end{document}